# Not All Models Suit Expert Offloading: On Local Routing Consistency of Mixture-of-Expert Models

**Jingcong Liang**
Fudan University
`jcliang22@m.fudan.edu.cn`

**Siyuan Wang**
University of Southern California
`sw_641@usc.edu`

**Miren Tian**
Huawei Technologies Ltd.
`tianmiren1@huawei.com`

**Yitong Li**
Huawei Technologies Ltd.
`liyitong3@huawei.com`

**Duyu Tang**
Huawei Technologies Ltd.
`tangduyu@huawei.com`

**Zhongyu Wei**
Fudan University
`zywei@fudan.edu.cn`

## Abstract

Mixture-of-Experts (MoE) enables efficient scaling of large language models (LLMs) with sparsely activated experts during inference. To effectively deploy large MoE models on memory-constrained devices, many systems introduce *expert offloading* that caches a subset of experts in fast memory, leaving others on slow memory to run on CPU or load on demand. While some research has exploited the locality of expert activations, where consecutive tokens activate similar experts, the degree of this **local routing consistency** varies across models and remains understudied. In this paper, we propose two metrics to measure local routing consistency of MoE models: (1) **Segment Routing Best Performance (SRP)**, which evaluates how well a fixed group of experts can cover the needs of a segment of tokens, and (2) **Segment Cache Best Hit Rate (SCH)**, which measures the optimal segment-level cache hit rate under a given cache size limit. We analyzed 20 MoE LLMs with diverse sizes and architectures and found that models that apply MoE on every layer and do not use shared experts exhibit the highest local routing consistency. We further showed that domain-specialized experts contribute more to routing consistency than vocabulary-specialized ones, and that most models can balance between cache effectiveness and efficiency with cache sizes approximately 2x the active experts. These findings pave the way for memory-efficient MoE design and deployment without compromising inference speed. We publish the code for replicating experiments at `https://github.com/ljcleo/moe-lrc`.

## 1 Introduction

Mixture-of-Experts (MoE) is a widely adopted model architecture in many large language models (LLMs) that enables efficient model size scaling through sparse activation[10, 17, 4, 1]. MoE models replace dense feed-forward networks (FFNs) with multiple expert modules, with only a subset activated during inference. However, the vanilla implementation requires all experts to be loaded into memory, restricting its application on memory-constrained devices such as mobile phones. To address this limitation, the *expert offloading* technique has been proposed to allow partial loading of expert modules during inference [8, 16, 53]. Specifically, expert offloading caches a subset of experts in fast memory (e.g., GPU memory) based on predefined heuristics, while storing remaining experts in slower but larger-capacity storage (e.g., CPU memory or disk). During inference, especially in the decoding stage, if a token activates an expert not cached in fast memory, the system either computes the expert forward results with CPU and slow memory (CPU offload) [18, 45], or unloads a cached expert according to specific rules such as Least Recently Used (LRU), and replaces it with the demanded expert (on-demand loading) [19, 57].

However, frequent CPU offloads or on-demand loading within a short period can significantly degrade the efficiency of the expert offloading system and slow down inference, particularly when processing lengthy contexts with inevitable topic shifts. Prior research has focused on optimizing the design of the expert offloading system, aiming to strategically select which experts to cache in a given context to maximize cache hit rates. Among them, some observed and exploited the locality of expert activations, where similar routing choices appear within a consecutive segment of tokens, thereby minimizing the need for CPU offloads and on-demand loading [8, 49, 56]. This is especially beneficial during the decoding phase, where tokens are generated one after another.

Nevertheless, not all MoE models exhibit such continuous routing patterns uniformly, and the degree or frequency of this phenomenon varies across models. Understanding this variance may help design MoE architectures that are friendly to expert offloading systems and vice versa. In this work, we investigate the degree of this inherent consecutive routing property, which we term **local routing consistency**, of different MoE-based LLMs to explore their potential effectiveness in segment-based expert routing or caching. Figure 1 illustrates how different levels of local routing consistency reflect different routing patterns. Specifically, we propose two metrics that quantitatively reflect the local routing consistency of a specific model. (1) **Segment Routing Best Performance (SRP)** measures how effectively a segment router selecting a fixed group of experts for all tokens in a segment can approximate the original router's decisions. SRP not only reflects local routing consistency *without parameters other than segment length*, but also enables analyzing activation patterns of *individual experts*. (2) **Segment Cache Best Hit Rate (SCH)** represents the highest cache hit rate by any expert offloading method that caches experts for whole segments, under a cache size limit related to the number of active experts. SCH measures local routing consistency on model- and router-levels, yet is more related to the performance of *real expert offloading systems* as it accounts for the cache limit.

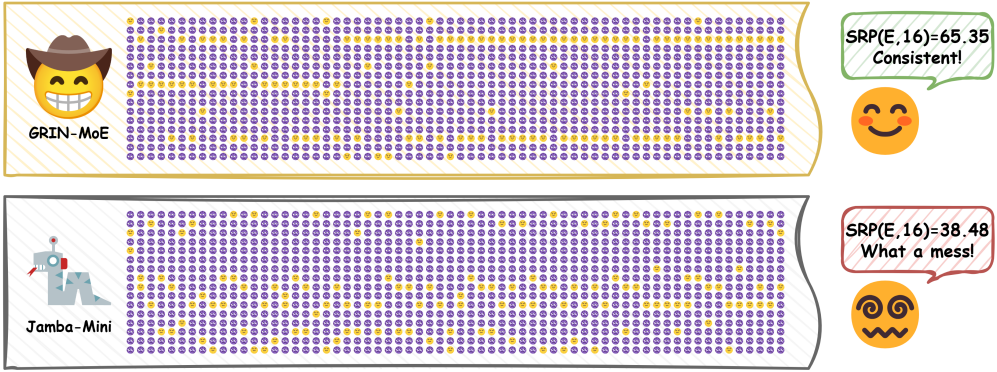

Figure 1: Routing results by GRIN-MoE [26] layer 21 and Jamba-Mini-1.6 [20] layer 25 on the same input (Java code). Despite having similar model sizes and the same number of experts, GRIN-MoE exhibits more consistent routing patterns than Jamba-Mini-1.6, with multiple experts continuously activated, so expert caching with GRIN-MoE will be more feasible and effective.

We conduct experiments on 20 MoE-based LLMs, covering parameter scales ranging from 3 billion to 54 billion and encompassing diverse architectures. While most models exhibit similar local routing consistency within a few tokens, the variance enlarges when the segment length increases. Models that **apply MoE on every layer** and **do not use shared experts** exhibit the **highest local routing consistency**, some of which even achieve load balance. Additionally, we investigate local routing consistency across different context domains and assess its relationship with experts' domain and vocabulary preferences. The findings reveal that **domain-specialized experts**, if they exist, **contribute more** to local routing consistency, whereas vocabulary specialization has less impact. Finally, we verify the strong correlation between SRP and SCH, confirming that both metrics effectively represent local routing consistency. We conclude that most MoE models can achieve **optimal balance** between segment caching effectiveness and deployment efficiency with a **cache size approximately 2x the number of active experts**.

Overall, our contributions are three-fold:

1. We propose *local routing consistency*, a property of MoE models that reflects the potential efficiency of expert offloading for the model. We design two metrics to quantify local routing consistency: segment routing best performance (SRP), which provides parameter-free fine-grained analysis, and segment cache best hit rate (SCH), aligned with practical expert offloading.
2. We conduct empirical analysis across 20 MoE-based LLMs, demonstrating that models that apply MoE on every layer and without shared experts show the highest local routing consistency. We also reveal that domain-specialized experts contribute more to local routing consistency than vocabulary-specialized experts.
3. We analyze SCH under different cache sizes relative to the number of active experts. Alongside the optimal cache size derived from SRP results, we conclude that cache sizes 2x the size of active parameters achieve the best segment caching results on most models.

## 2 Definitions

### 2.1 Preliminary: mixture of experts

Transformer-based language models have most parameters on feed-forward network (FFN) layers. As the model size scales up, LLMs often replace them with sparse MoE layers to reduce computational cost during inference. A typical MoE layer with $E$ experts can be parameterized by $E$ smaller FFNs $F_1(;\theta_1), \ldots, F_E(;\theta_E)$, where each $F_i \colon \mathbb{R}^d \to \mathbb{R}^d$ defines a single expert. There is also a router $R \colon \mathbb{R}^d \to \mathbb{R}^E$ in the MoE layer to choose experts and give weights. For each token $x$ with hidden representation $h_x \in \mathbb{R}^d$, its output is given by

$$[s_1, \ldots, s_E] = \mathrm{Softmax}(R(h_x)); \ [w_1, \ldots, w_E] = \mathrm{Top}_k(s_1, \ldots, s_E); \ o_x = \sum_{i=1}^{E} w_i F_i(h_x; \theta_i) \quad (1)$$

where $\mathrm{Top}_k$ preserves the $k$ largest scores and sets others to $0$. Experts with a null score can be effectively deactivated without calculating, thus saving computational resources. Other components in the transformer architecture may also be replaced by their MoE variant, such as Mixture-of-Attention [55] for self-attention and MixLoRA [22] for LoRA adapters. Nevertheless, we focus on MoE layers that replace FFNs, as this is the most prominent and effective design (in terms of the number of parameters). More discussions about MoE LLMs can be found in Appendix A, where we also briefly review expert offloading for MoE models.

The above routing procedure is done token by token, which does not guarantee consecutive routing. Since the consistency of consecutive routing decisions can benefit expert offloading systems, it is necessary to investigate the degree of consecutive routing of different MoE-based LLMs. In the following sections, we propose two metrics to measure this **local routing consistency** and compare them across MoE models with different structure parameters.

### 2.2 Segment routing best performance (SRP)

An intuitive way to measure local routing consistency is to compare the distribution of routing choices between tokens in a continuous segment. However, typical metrics for distribution comparison, such as the Kullback-Leibler divergence, work with two distributions and thus are unsuitable for our demand. Instead, we measure how well a simplified, segment-based router can mimic the behavior of the original token-based router. Here we define the **segment routing best performance (SRP)** of a single expert $e$ and a group of experts $E$ (e.g., experts from the same MoE layer or model).

**Single expert** For any input sequence $T = [t_1, \ldots, t_{|T|}]$, we denote the activation sequence of expert $e$ on $T$ as $A(e, T) = [a_1, \ldots, a_{|T|}]$, where $a_i \in \{0, 1\}$ indicates whether $e$ is activated on $t_i$ ($a_i = 1$) or not ($a_i = 0$). A segment-based router $R_e^m$ with segment length $m > 0$ for expert $e$ will try to mimic $[a_p, \ldots, a_{p+m-1}]$ for any $T$ and $p$, similar to expert-choice routing [59]. Let $R_e^m(T, p) = [b_1, \ldots, b_m]$ be its prediction, with the segment-prediction constraint:

$$b_i = 0, \ \forall i = 1, \ldots, m \quad \text{or} \quad b_i = 1, \ \forall i = 1, \ldots, m \quad (2)$$

In other words, $R_e^m$ decides that $e$ is either always active or always inactive on $[t_p, \ldots, t_{p+m-1}]$. For simplicity, we write $R_e^m(T, p) = 0$ and $R_e^m(T, p) = 1$ for the two cases respectively. By treating each segment routing attempt as a binary classification task with $m$ samples, and considering all possible segments of all possible inputs, we can calculate the $F_1$ score of $R_e^m$:

$$F_1(R_e^m) = \frac{2 \sum_T \sum_{p=1}^{|T|-m+1} R_e^m(T,p) \cdot f(e,T,p,m)}{\sum_T \sum_{p=1}^{|T|-m+1} [m \cdot R_e^m(T,p) + f(e,T,p,m)]} \quad (3)$$

where $f(e,T,p,m) = \sum_{i=p}^{p+m-1} A(e,T)[i]$ is the active frequency of $e$ in the segment of $T$ with length $m$ starting at position $p$. We demonstrate the detailed process to obtain this equation in Appendix B.1. We choose $F_1$ score because we do not set up an upper bound on how many experts $R_e^m$ can be selected, so hit rate (recall) can go up to 1 by selecting all experts. Moreover, missing major experts is worse than activating minor ones, as changing low-rank experts usually affects model performance less [19, 43], therefore we prefer $F_1$ score to accuracy to emphasize this difference.

Based on Equation 3, we define the segment routing best performance of $e$ under segment length $m$ as the maximum $F_1$ score any $R_e^m$ can achieve: $\mathrm{SRP}(e,m) \triangleq \max_{R_e^m} F_1(R_e^m)$. Furthermore, in Appendix B.2 we prove that $F_1(R_e^m)$ is maximized if and only if $R_e^m$ gives active predictions for all segments that activates $e$ at least $\alpha_e^m$ times, where $\alpha_e^m \in [0,m]$ is only related to $e$ and $m$:

$$\mathrm{SRP}(e,m) = \frac{2 \sum_T \sum_{f(e,T,p,m) \geq \alpha_e^m} f(e,T,p,m)}{\sum_T \sum_{p=1}^{|T|-m+1} [m \cdot I[f(e,T,p,m) \geq \alpha_e^m] + f(e,T,p,m)]} \quad (4)$$

Therefore, $\mathrm{SRP}(e,m)$ is an intrinsic property of the expert that reflects its local routing consistency, unrelated to any specific segment routing methods.

**Expert group** For a group of experts $E$, let $R_E^m$ be a segment-based router that decides whether each expert $e \in E$ should be activated in a segment of some input $T$ with length $m$; more specifically, $R_E^m(e,T,p)$ is a prediction sequence similar to $R_e^m(T,p)$ that also follows the segment-prediction constraint (Equation 2). Following the same procedure in Appendix B.1, we have

$$F_1(R_E^m) = \frac{2 \sum_T \sum_{p=1}^{|T|-m+1} \sum_{e \in E} R_E^m(e,T,p) \cdot f(e,T,p,m)}{\sum_T \sum_{p=1}^{|T|-m+1} \sum_{e \in E} [m \cdot R_E^m(e,T,p) + f(e,T,p,m)]} \quad (5)$$

Again, $F_1(R_E^m)$ is maximized if and only if $R_E^m$ gives active predictions for all expert-segment pairs where the expert is activated at least $\alpha_e^m$ times in the segment, here $\alpha_e^m$ decided by $E$ and $m$. Therefore we have

$$\mathrm{SRP}(E,m) \triangleq \max_{R_E^m} F_1(R_E^m) = \frac{2 \sum_T \sum_{f(e,T,p,m) \geq \alpha_e^m} f(e,T,p,m)}{\sum_T \sum_{p=1}^{|T|-m+1} \sum_{e \in E} [m \cdot I[f(e,T,p,m) \geq \alpha_e^m] + f(e,T,p,m)]} \quad (6)$$

$\mathrm{SRP}(E,m)$ measures how well a group of experts is coordinated by the original router(s) to achieve layer-level or model-level local routing consistency.

**Segment routing size ratio** Consider the case where the expert group $E$ is the set of all experts from an MoE layer that activates top-$k$ experts. When we use the prediction of $R_E^m$ that achieves the highest $F_1$ to route experts, for any segment of input $T$ with length $m$ starting at position $p$, only experts that is activated at least $\alpha_e^m$ times by the original router will be chosen. Hence, the expectation of the number of routed experts for any segment is given by

$$k_m = \sum_T \frac{1}{|T|-m+1} \sum_{p=1}^{|T|-m+1} \sum_{e \in E} I[f(e,T,p,m) \geq \alpha_e^m] \quad (7)$$

That is, $R_E^m$ will select $k_m$ experts on average. We define the ratio between $k_m$ and $k$ as the **segment routing size ratio**: $\rho(E,m) \triangleq k_m/k$. A small $\rho(E,m)$ indicates that the local routing consistency of the experts is high enough, so that segment routing does not need to select too many experts to cover real demands under average cases. We use it as a supplementary metric to distinguish cases when groups of experts have similar segment routing best performances. Note that $\rho(E,m)$ is also available for the entire model if the total number of activated experts for each token is also fixed.

Note that in the above definitions, $T$ is unbounded, causing $\mathrm{SRP}(e, m)$, $\mathrm{SRP}(E, m)$ and $\rho(E, m)$ to be uncomputable. In experiments, we choose a fixed corpora $S$ (general or domain-specific) and let $T \in S$, in which case we write $\mathrm{SRP}_S(e, m)$, $\mathrm{SRP}_S(E, m)$ and $\rho_S(E, m)$ to indicate the corpora. Furthermore, we only need to consider $\alpha_e^m$ and $\alpha_E^m$ that are integers, thus we can effectively enumerate from 0 to $m$ to find the corresponding $\alpha_e^m$ and $\alpha_E^m$ respectively.

## 2.3 Segment cache best hit rate (SCH)

The advantage of SRP as a local routing consistency metric is that it only relies on the expert $e$ (or expert group $E$) and the segment length $m$, and is suitable to analyze individual experts. However, real expert offloading scenarios usually have a hard cache size limit, so the best global $F_1$ score may not be achievable. Moreover, when considering caching performance, $F_1$ score is not as straightforward as hit rate (recall). Therefore, we propose another metric for local routing consistency, namely **segment cache best hit rate (SCH)**, that is more related to expert offloading.

We use the same notations as above, and stick to the expert group case where $E$ denotes a group of experts that activates $k$ experts for each token. Instead of a segment-based router $R_E^m$, here we consider a segment-based cache $C_E^{m,\rho}$ where $m$ is the segment length, and $\rho$ be the **segment cache size ratio**; in other words, $C_E^{m,\rho}$ has a cache size of $\rho k$. For any input $T$ and any of its sub-segment of length $m$ starting at position $p$, $C_E^{m,\rho}$ will choose $\rho k$ experts to cache throughout the segment, without switching any experts. Let $C_E^{m,\rho}(e, T, p) = 1$ if an expert $e \in E$ is cached for the segment, and $C_E^{m,\rho}(e, T, p) = 0$ otherwise, then the hit rate of $C_E^{m,\rho}$ across all segments of all possible inputs is given by

$$r(C_E^{m,\rho}) = \frac{\sum_T \sum_{p=1}^{|T|-m+1} \sum_{e \in E} C_E^{m,\rho}(e, T, p) \cdot f(e, T, p, m)}{\sum_T \sum_{p=1}^{|T|-m+1} \sum_{e \in E} f(e, T, p, m)} \tag{8}$$

We define the segment cache best hit rate of $E$ under segment length $m$ and segment cache size ratio $\rho$ as the maximum hit rate any $C_E^{m,\rho}$ can reach: $\mathrm{SCH}(E, m, \rho) = \max_{C_E^{m,\rho}} r(C_E^{m,\rho})$. Note that in Equation 8, the denominator is unrelated to $C_E^{m,\rho}$; therefore, to maximize $r(C_E^{m,\rho})$, we only need to make sure that $C_E^{m,\rho}$ always caches the $\rho k$ experts with the highest $f$:

$$\mathrm{SCH}(E, m, \rho) = \frac{\sum_T \sum_{p=1}^{|T|-m+1} \sum \mathrm{Top}_{\rho k}\{f(e, T, p, m) | e \in E\}}{\sum_T \sum_{p=1}^{|T|-m+1} \sum_{e \in E} f(e, T, p, m)} \tag{9}$$

We write $\mathrm{SCH}_S(E, m, \rho)$ when we limit $T$ to a fixed corpora $S$.

# 3 SRP-based consistency analysis

## 3.1 Experiment setup

**Models** We conduct experiments on 20 MoE-based LLMs with various model sizes:

- 0∼10B: PowerMoE-3B [42], LLaMA-MoE-v1-3.5B [60], OLMoE-1B-7B-0125 [29], Switch-Transformers-Base-128 [10], LLaMA-MoE-v2-3.8B [33] and JetMoE-8B [41];
- 10∼25B: OpenMoE-8B [48], MiniCPM-MoE-8x2B [13], Qwen1.5-MoE-A2.7B [34], DeepSeek-V2-Lite [5] and DeepSeekMoE [4];
- 25∼45B: XVERSE-MoE-A4.2B [50], Qwen3-30B-A3B [35], Yuan2.0-M32 [47], Phi-3.5-MoE [1] and GRIN-MoE [26];
- 45∼60B: Mixtral-8x7B-v0.1 [17], Jamba-Mini-1.6 [20], NLLB-MoE-54B [31] and Qwen2-57B-A14B [51].

We list their architecture and configuration details in Appendix C.1, and may use shorter names (e.g., OLMoE and Qwen3) when there is no ambiguity.

**Data** We construct our sample corpus $S$ from RedPajama [46], including 7 data sources: C4, CommonCrawl, Books, Wikipedia, ArXiv, StackExchange, and GitHub, each representing a specific domain. Appendix C.2 introduces data processing and input generation details. In the following sections, we denote the full corpus as $S$, and each domain's corresponding subset as its data source name (e.g., Books, GitHub).

**Method and configuration** We collect every MoE layer's routing decisions for every input[*], and for each expert, count the number of activated tokens $f$ in every segment. To obtain the SRP, we count the number of segments with the same $f$ for each expert (group), then compute the $F_1$ score for every $\alpha$ candidate, choose the $\alpha$ that achieves the highest $F_1$ and finally obtain the segment routing best performance and size ratio. We run all experiments on 8 NVIDIA GeForce RTX 4090 graphics cards and load model parameters with BF16 precision.

## 3.2 Overall results

Figure 2 illustrates all models' SRP. While most models have similar SRP and $\rho$ when $m = 4$, the difference between models becomes significant as the segment length increases. Meanwhile, the relative distribution of SRP across models appears similar after $m = 16$. There is a gap between short-term ($m = 4$) and long-term ($m \geq 16$) local routing consistency, where **many models exhibit the short-term one but only a few demonstrate the long-term one**.

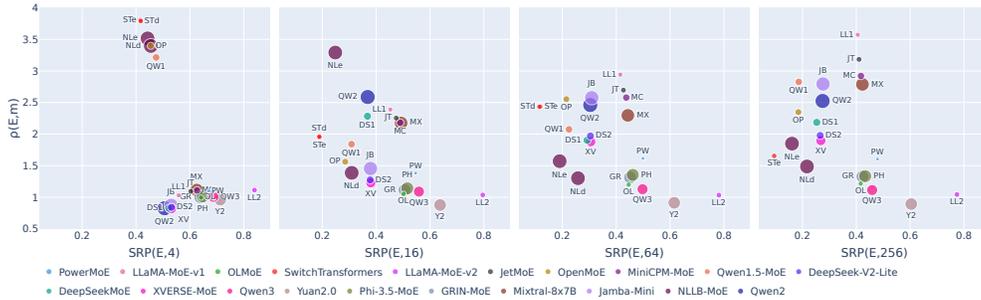

Figure 2: Model SRP on $S$, compared with the segment routing size ratio. Marker size represents model size.

We roughly divide the models into four groups that have similar SRP characteristics, whose SRP when $m = 16$ are demonstrated in Table 1:

- Group 1 (LLaMA-MoE-v2–OLMoE) has the highest SRP ($> 0.5$ when $m = 16$) and $\rho$ ($\sim 1.25$) across all segment lengths, showing strong long-term local routing consistency.
- Group 2 (Mixtral-8x7B–LLaMA-MoE-v1) have the second highest SRP ($\sim 0.48$ when $m = 16$), but their long-term $\rho$ becomes high ($\sim 2.5$).
- Group 3 (XVERSE-MoE–DeepSeekMoE) has a significantly lower SRP than group 2, especially the long-term one ($\sim 0.36$ when $m = 16$), but their $\rho$ is a bit lower $\sim 2.0$.
- Group 4 (NLLB-MoE–SwitchTransformers) have the lowest SRP ($< 0.31$ when $m = 16$); contrasting other models, their short-term $\rho$ is already high, but their long-term $\rho$ becomes lower.

Table 1: Model segment routing best performance ($m = 16$) on $S$ in descending order and several architecture parameters. "A:T": ratio between active and all experts; "S:A": ratio between shared and active experts; "every $x$": apply MoE every $x$ layers; "after 1st": apply MoE after the first layer.

| Model | SRP | MoE | A:T | S:A | Model | SRP | MoE | A:T | S:A |
|---|---|---|---|---|---|---|---|---|---|
| LLaMA-MoE-v2 | 79.59 | all | 1:4 | 0 | XVERSE-MoE | 38.03 | all | 3:32 | 1:3 |
| Yuan2.0 | 63.75 | all | 1:16 | 0 | Jamba-Mini | 37.91 | every 2 | 1:8 | 0 |
| Qwen3 | 55.90 | all | 1:16 | 0 | DeepSeek-V2-Lite | 37.79 | after 1st | 3:32 | 1:3 |
| PowerMoE | 54.69 | all | 1:5 | 0 | Qwen2 | 36.86 | all | 1:8 | 1:1 |
| Phi-3.5-MoE | 51.59 | all | 1:8 | 0 | DeepSeekMoE | 36.79 | after 1st | 3:32 | 1:3 |
| GRIN-MoE | 50.55 | all | 1:8 | 0 | | | | | |
| OLMoE | 50.17 | all | 1:8 | 0 | NLLB-MoE (encoder) | 24.84 | every 4 | 1:64 | 0 |
| | | | | | (decoder) | 30.94 | | | |
| Mixtral-8x7B | 49.27 | all | 1:4 | 0 | Qwen1.5-MoE | 30.88 | all | 1:15 | 1:1 |
| MiniCPM-MoE | 48.93 | all | 1:4 | 0 | OpenMoE | 28.54 | every 6 | 1:16 | 1:2 |
| JetMoE | 47.46 | all | 1:4 | 0 | SwitchTF (encoder) | 18.98 | every 2 | 1:128 | 0 |
| LLaMA-MoE-v1 | 45.28 | all | 1:4 | 0 | (decoder) | 18.90 | | | |

---

[*]In encoder-decoder models, encoder layers only consider encoder input, and decoder layers likewise.

### 3.3 Analysis

**Which model architecture has the highest local routing consistency?** We compare the SRP of different models with some architecture parameters in Table 1. Surprisingly, the least sparse models **do not always** have very high local routing consistency, even though they activate experts more frequently on average. Instead, we observe two common points that are possibly connected to high local routing consistency: (1) **Apply MoE on every layer:** all models in groups 1 and 2 apply the MoE structure on every layer, while many group 3 and 4 models skip some layers (e.g., Jamba). (2) **No shared experts:** all models in groups 1 and 2 do not include any shared experts, unlike many group 3 and 4 models (e.g., Qwen2). We conjecture that in this way, models gain higher local routing consistency by fully utilizing MoE. We also compare SRP with corpus perplexity (Appendix D.2) and other model design aspects (Appendix D.3), where we find little correlation.

Table 2: Model segment routing best performance ($m = 16$) on $S$, along with load balance (LB) measured by activation frequency standard deviation of experts.

| Model | SRP | LB | Model | SRP | LB | Model | SRP | LB |
|---|---|---|---|---|---|---|---|---|
| LLaMA-MoE-v2 | 79.59 | 30.46 | Mixtral-8x7B | 49.27 | 2.53 | Qwen2 | 36.86 | 6.72 |
| Yuan2.0 | 63.75 | 13.94 | MiniCPM-MoE | 48.93 | 2.81 | DeepSeekMoE | 36.81 | 1.56 |
| Qwen3 | 55.90 | 4.35 | JetMoE | 47.46 | 0.99 | NLLB-MoE (en) | 24.84 | 1.67 |
| PowerMoE | 54.69 | 12.39 | LLaMA-MoE-v1 | 45.28 | 2.67 | (de) | 30.94 | 2.06 |
| Phi-3.5-MoE | 51.59 | 4.06 | XVERSE-MoE | 38.03 | 2.39 | Qwen1.5-MoE | 30.88 | 0.55 |
| GRIN-MoE | 50.55 | 3.48 | Jamba-Mini | 37.91 | 2.91 | OpenMoE | 28.54 | 2.59 |
| OLMoE | 50.17 | 5.24 | DeepSeek-V2-Lite | 37.79 | 1.93 | SwitchTF (en) | 18.98 | 0.48 |
| | | | | | | (de) | 18.90 | 0.62 |

**Can local routing consistency live with load balance?** Load balance is a key feature for efficient MoE inference [21], but is seemingly against local routing consistency. For instance, both DeepSeek-AI et al. [6] and Skliar et al. [43] suggest adding bias to router outputs. Still, the former promotes *little activated* experts for load balance and the latter promotes *recently cached* experts for effective caching. To investigate their relation, we compute the standard deviation of all experts' activation frequencies in a model and compare it with SRP in Table 2. Many models with high SRP also have high activation frequency standard deviation. Appendix D.8 reveals that their local routing consistency, at least partly, comes from expert imbalance. However, models like Qwen3 and GRIN-MoE enjoy high local routing consistency and moderate load balance simultaneously, in which we found strong domain-specialized experts (see Section 4). Based on the observations, we claim that **local routing consistency can coexist with load balance through domain-specialized experts.**

## 4 Local routing consistency and expert specialization

### 4.1 Domain-wise local routing consistency

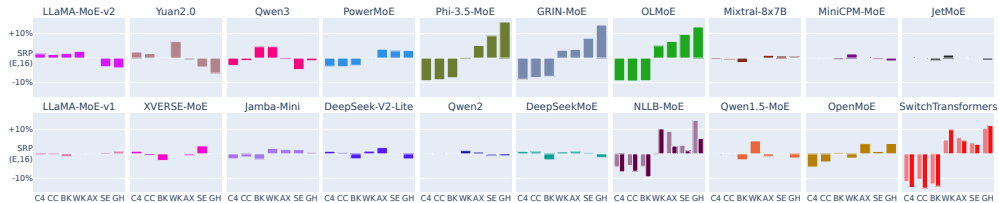

Figure 3: Segment routing best performance ($m = 16$) on each domain, relative to $\mathrm{SRP}_S(E, 16)$. CC: CommonCrawl; BK: Books; WK: Wikipedia; AX: ArXiv; SE: StackExchange; GH: GitHub. For encoder-decoder models, light color represents the encoder and dark color represents the decoder.

In Section 3, we analyze models on the full corpus $S$, which consists of text data from 7 different domains. However, each domain has its token distributions, which may affect the router decision's distribution. Figure 3 illustrates the relative difference between domain-wise and global SRP of each model when $m = 16$, where we observe three different patterns among all models: (1) Models like Phi-3.5-MoE, GRIN-MoE, and OLMoE have significantly **higher SRP on Wikipedia, ArXiv, StackExchange and GitHub**, whose SRP can be more than 10% higher than global SRP. All these domains are *professional or related to code*. (2) Models like Yuan2.0, Qwen3 and Qwen1.5-MoE have significant **higher SRP on Wikipedia**, but **not on other three professional domains**; compared with other domains, Wikipedia contains *multilingual text*, which may be the cause of such uniqueness. (3) Models like Mixtral-8x7B, MiniCPM-MoE and JetMoE have **similar SRP across all domains** with insignificant differences. Most of these models have *mediocre to low SRP*. Above all, models exhibit balanced local routing consistency across all domains or higher local routing consistency on certain domains with unique properties.

## 4.2 Expert specialization

We argue that domain-wise local routing consistency patterns appear across models due to specialized experts in each model. To clarify this, we consider two types of expert specialization, first introduced by Muennighoff et al. [29]: (1) **Domain specialization:** the normalized frequency of an expert $e$ being activated on tokens from a specific domain $D$. We compute the coefficient of variation (CV) of activation frequency across all domains as a domain-free metric. (2) **Vocabulary specialization:** the normalized frequency of an expert $e$ being activated on a specific token ID $x$. We follow Muennighoff et al. [29] to obtain the vocabulary specialization of each expert. We compare each model's SRP, average expert specialization, and the correlation between experts' specialization and SRP in Figure 4; expert distribution between specialization and SRP is also demonstrated in Appendix D.8.

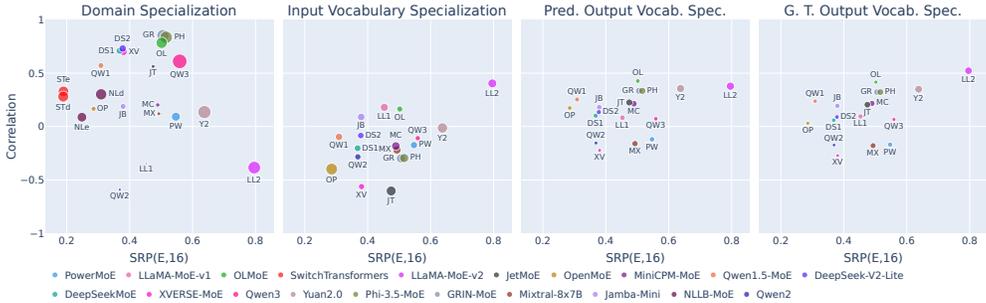

Figure 4: Model segment routing best performance ($m = 16$), compared with average expert specialization (marker size) and the correlation between expert specialization and SRP (y-axis).

**Domain specialization** Most models show a positive correlation between their experts' domain specialization and SRP; among them, Qwen3, Phi-3.5-MoE, GRIN-MoE, and OLMoE show high SRP, high average domain specialization, and strong correlation between them simultaneously, indicating that these models possess **many domain specialized experts** that **make major contribution to local routing consistency**. An exception is LLaMA-MoE-v2, which is very imbalanced: it constantly activates a group of experts, resulting in high SRP and low domain specialization. Nevertheless, both groups of experts contribute to local routing consistency, giving the model the highest SRP.

**Vocabulary specialization** We consider three kinds of vocabulary specialization on the input, the model's predicted output, and the ground-truth, respectively [29]. Most models demonstrate **negative or insignificant correlation between input vocabulary specialization and SRP**; LLaMA-MoE-v2 becomes the only exception, also due to the constantly activated experts. On the other hand, **SRP is slightly positively correlated to prediction or ground truth vocabulary specialization.** We conjecture that such specialization happens more in later layers [29] that process high-level information related to the context topic.

Above all, we can see that **domain specialization plays a more important role in forming local routing consistency than vocabulary specialization**, especially on load-balanced models.

## 5 SCH-based consistency analysis

### 5.1 Overall results

As mentioned in Section 2.3, SRP has several flaws that hinder its application in expert offloading. This section focuses on the segment cache best hit rate (SCH), which works with a size limit, to obtain a more straightforward insight into expert offloading and cache management. Based on the number of activated tokens in every input segment from the corpus for each expert, defined as $f$ in Section 3.1, we calculate each model and each layer's SCH on every possible cache size: At each segment, we sort experts in each layer or model by their $f$ in descending order, and compute the segment cache best hit rate at cache sizes ranging from 1 to $|E|$. Figure 5 illustrates $\mathrm{SCH}(E, m, \rho$ of each model under different $m$s and $\rho$s. Similar to SRP, we can easily locate the four groups of models mentioned in Section 3.2 starting from $m = 16$. We can also identify the minimum $\rho$ of each group for some specific cache hit rate threshold (e.g., 0.6):

- Group 1 models have $\mathrm{SCH}(E, m, 2) \geq 0.6$ for segment lengths $m$ up to 256. They also have turning points between $\rho = 1.5$ and $\rho = 2$, after which increasing $\rho$ improves SCH less.
- Group 2 models also have $\mathrm{SCH}(E, m, 2) \geq 0.6$ as $m$ increases, but their turning points appear at $\rho > 2$ and are not significant, since these models are not sparse enough.
- For group 3 models, $\rho = 2$ is sufficient or reach a segment cache hit rate around 0.6 when $m = 16$, but quickly becomes not enough as $m$ increases.
- Group 4 models have the slowest growth rate of SCH and require $\rho \geq 3$ to make $\mathrm{SCH}(E, 16, \rho)$ reach 0.6. Models in groups 3 and 4 also lack turning points at $\rho \leq 4$ as group 1 models do.

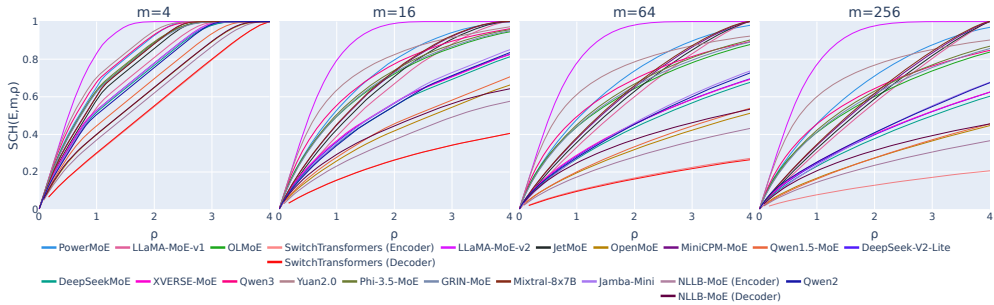

Figure 5: Model SCH score under different segment length $m$ and segment cache size ratio $\rho$.

Above all, increasing $\rho$ from $\rho = 2$ improves SCH less for group 1 models, yet decreasing $\rho$ can harm caching performance for group 2 and 3 models. As recent MoE LLMs fall into the first three groups, we claim that in general, $\rho = \mathbf{2}$ **can balance cache effectiveness and efficiency.**

### 5.2 Relation with segment routing best performance

To clarify the relation between $\mathrm{SCH}(E, m, \rho)$ and $\mathrm{SRP}(E, m)$, Table 3 lists the correlation between them across all models. $\mathrm{SRP}(E, m)$ and $\mathrm{SCH}(E, m, \rho)$ are always highly positively correlated regardless of the values of $m$ and $\rho$. This ensures that SCH shares the same property of SRP under reasonable segment length and cache size. Furthermore, when $\rho$ is around 1.5, the two metrics are most closely related, nearly perfectly linear, aligned with Figure 2 where most models have $\rho(E, m)$ between 1–3 when $m \geq 16$, as well as our previous claim that $\rho = 2$ balances cache effectiveness and efficiency.

Table 3: Correlation between $\mathrm{SCH}(E, m, \rho)$ and $\mathrm{SRP}(E, m)$ across all models. Bold font indicates the highest correlation across $\rho$ for each $m$.

| $\rho$ | $m = 4$ | $m = 16$ | $m = 64$ | $m = 256$ |
|---|---|---|---|---|
| 0.5 | 96.58 | 97.95 | 96.86 | 95.57 |
| 1.0 | 97.98 | 99.54 | 99.04 | 98.36 |
| 1.5 | **98.15** | **99.73** | **99.77** | **99.57** |
| 2.0 | 95.76 | 98.32 | 98.90 | 99.08 |
| 2.5 | 88.58 | 95.33 | 96.49 | 96.80 |
| 3.0 | 74.44 | 91.86 | 93.13 | 93.20 |

# 6 Conclusion

In this paper, we investigate the property of MoE LLMs where similar experts can be continuously activated, namely *local routing consistency*. We propose two metrics to measure this property: segment routing best performance (SRP) and segment cache best hit rate (SCH). We found several key designs that may help improve local routing consistency of MoE LLMs, and suggest that a cache size approximately 2x the number of active experts can balance cache effectiveness and efficiency.

**Broader Impacts and Limitations** The primary limitations of our work are the computational constraints restricting MoE LLMs to under 60B parameters and the theoretical nature of SRP and SCH, which focus on pure segment routing/caching. Future work will investigate larger models like DeepSeek-V3 and validate the applicability of our findings and claims to real expert offloading systems. Nonetheless, our analytic methods and results still help design new MoE LLMs friendly to expert offloading and enable deployment on resource-constrained edge devices. While our study will likely have an indirect social impact, developers implementing local routing consistency to build more powerful LLMs must take responsibility for their products' societal implications.

## A Related work

### A.1 MoE-based LLM and expert analysis

Since its introduction into large neural networks, MoE has become a critical strategy to build large language models up to trillions of parameters [38, 10, 36]. While some early models like SwitchTransformers [10] and NLLB [31] employ encoder-decoder structures as their backbone, due to the success of GPT-3, most recent popular MoE-based LLMs use decoder-only structures [17, 51, 6, 1], replacing their original FFN layers with MoE layers containing multiple experts (other components may be replaced too, e.g., self-attention [40, 41] and LoRA [22, 11]). Cai et al. [2] systematically introduces MoE architectures and implementation in LLMs.

The popularity of MoE LLMs has triggered interest in understanding how experts are activated in such models. Many model reports and individual studies focused on the relation between expert selection and the input context. For example, Muennighoff et al. [29] reported that OLMoE shows a significant difference in expert activity across different domains. Contrastively, Xue et al. [48]

found that the routing choice of OpenMoE is highly related to the input token rather than the input context. Other works investigate the similarity among expert activation patterns [24, 28], as well as the relation between expert output and routing choice [32, 27]. However, few of them have focused on the local activation pattern of experts. For example, Jiang et al. [17] reported that in Mixtral, experts are more likely to be activated consecutively, compared to the random case. While their results provide fundamental support for many efficient MoE inference systems Liu et al. [25], they only examined the case of 2 consecutive tokens, which may be insufficient to ensure the consistency of expert activation in longer segments.

### A.2   Efficient MoE inference and expert offloading

The discrete nature of vanilla MoE routers and redundant parameters has caused MoE models to infer more slowly and consume more memory than dense models with the same number of activated parameters (thus with the same level of FLOPs). Many techniques have been proposed to boost the inference of MoE models, ranging from model modifications like model compression [3, 15, 52, 36] and soft routing [30, 58] to system implementations like load-balanced expert parallel [21, 14, 23] and hardware adaptation [6, 53]. Liu et al. [25] provides an in-depth summary of various inference optimization strategies of MoE models.

In this paper, we mainly focus on the potential performance of expert offloading, which enables lossless inference of MoE models on memory-constrained devices by caching only some experts on (fast) memory while leaving others on slow memory or disk storage. Many such systems use pretrained external models and/or information from previous layers to prefetch experts for upcoming layers [37, 7, 12, 44]. Various expert offloading systems propose curated heuristics to manage the expert cache [43, 49, 54, 9]. Among them, some examine the locality of expert activations as empirical support for expert caching efficiency. For example, Eliseev and Mazur [8] found that Mixtral-8x7B-Instruct can activate the same expert across 2-4 consecutive tokens, similar to the observations by Jiang et al. [17]; Xue et al. [49] reported frequent expert reuse during decoding, and Zhang et al. [56] found similar routing choice between prefilling and decoding stages. However, most previous works focus on models like SwitchTransformers and Mixtral; only a few have included more recent MoE LLMs. As more MoE LLMs emerge, understanding what models are more friendly to expert offloading becomes important for the development of both MoE architectures and expert offloading methods.

## B   Proofs

### B.1   Proof of equation 3

In Section 2.2, we consider each routing decision of $R_e^m$ for a segment of length $m$ as a binary classification task with $m$ samples. If we merge all samples from all segments of all possible inputs into one global binary classification task, and define $f(e, T, p, m) = \cdot \sum_{i=p}^{p+m-1} A(e, T)[i]$ as in

Section 2.2, we will have the following prediction statistics:

$$\text{TP}(\text{R}_\text{e}^\text{m}) = \sum_T \sum_{p=1}^{|T|-m+1} \sum_{i=1}^{m} I[A(e,T)[p+i-1] = 1 \wedge R_e^m(T,p)[i] = 1]$$

$$= \sum_T \sum_{p=1}^{|T|-m+1} R_e^m(T,p) \cdot f(e,T,p,m) \tag{10}$$

$$\text{FP}(\text{R}_\text{e}^\text{m}) = \sum_T \sum_{p=1}^{|T|-m+1} \sum_{i=1}^{m} I[A(e,T)[p+i-1] = 0 \wedge R_e^m(T,p)[i] = 1]$$

$$= \sum_T \sum_{p=1}^{|T|-m+1} R_e^m(T,p)[m - f(e,T,p,m)] \tag{11}$$

$$\text{FN}(\text{R}_\text{e}^\text{m}) = \sum_T \sum_{p=1}^{|T|-m+1} \sum_{i=1}^{m} I[A(e,T)[p+i-1] = 1 \wedge R_e^m(T,p)[i] = 0]$$

$$= \sum_T \sum_{p=1}^{|T|-m+1} [1 - R_e^m(T,p)] f(e,T,p,m) \tag{12}$$

Therefore we have

$$\begin{aligned}
F_1(R_e^m) &= \frac{1}{1/\text{Precision}(R_e^m) + 1/\text{Recall}(R_e^m)} \\
&= \frac{1}{[TP(R_e^m) + FP(R_e^m)]/TP(R_e^m) + [TP(R_e^m) + FN(R_e^m)]/TP(R_e^m)} \\
&= \frac{2TP(R_e^m)}{[TP(R_e^m) + FP(R_e^m)] + [TP(R_e^m) + FN(R_e^m)]} \\
&= \frac{2 \sum_T \sum_{p=1}^{|T|-m+1} R_e^m(T,p) \cdot f(e,T,p,m)}{\left[\sum_T \sum_{p=1}^{|T|-m+1} m \cdot R_e^m(T,p)\right] + \left[\sum_T \sum_{p=1}^{|T|-m+1} f(e,T,p,m)\right]} \\
&= \frac{2 \sum_T \sum_{p=1}^{|T|-m+1} R_e^m(T,p) \cdot f(e,T,p,m)}{\sum_T \sum_{p=1}^{|T|-m+1} [m \cdot R_e^m(T,p) + f(e,T,p,m)]}
\end{aligned} \tag{13}$$

which gives Equation 3. $\square$

## B.2 Proof of equation 4

Assume that $R_e^m(T_0, p_0) = 0$ for some $e, m > 0, R_e^m, T_0$ and $p_0$, then we have

$$\begin{aligned}
F_1(R_e^m) &= \frac{2 \sum_T \sum_{p=1}^{|T|-m+1} R_e^m(T,p) \cdot f(e,T,p,m)}{\sum_T \sum_{p=1}^{|T|-m+1} [m \cdot R_e^m(T,p) + f(e,T,p,m)]} \\
&= \frac{2 \sum_T \sum_{p=1}^{|T|-m+1} R_e^m(T,p) \cdot f(e,T,p,m)}{m \sum_T \sum_{p=1}^{|T|-m+1} R_e^m(T,p) + \sum_T \sum_{p=1}^{|T|-m+1} f(e,T,p,m)} \\
&= \frac{2 \sum_{T \neq T_0 \wedge p \neq p_0} R_e^m(T,p) \cdot f(e,T,p,m)}{m \sum_{T \neq T_0 \wedge p \neq p_0} R_e^m(T,p) + \sum_T \sum_{p=1}^{|T|-m+1} f(e,T,p,m)} \\
&= \frac{2X}{mY + Z}
\end{aligned} \tag{14}$$

where

$$X = \sum_{\substack{T \neq T_0 \\ p \neq p_0}} R_e^m(T,p) \cdot f(e,T,p,m), \quad Y = \sum_{\substack{T \neq T_0 \\ p \neq p_0}} R_e^m(T,p), \quad Z = \sum_T \sum_{p=1}^{|T|-m+1} f(e,T,p,m)$$

Let $\widehat{R_e^m}$ be a copy of $R_e^m$ except that $R_e^m(T, p) = 1$; all other routing decisions remain the same. Then the $F_1$ score of the new segment router will be

$$
\begin{aligned}
F_1\left(\widehat{R_e^m}\right) &= \frac{2\sum_T \sum_{p=1}^{|T|-m+1} \widehat{R_e^m}(T, p) \cdot f(e, T, p, m)}{\sum_T \sum_{p=1}^{|T|-m+1}\left[m \cdot \widehat{R_e^m}(T, p) + f(e, T, p, m)\right]} \\
&= \frac{2[X + f(e, T_0, p_0, m)]}{m(Y + 1) + Z} \\
&= \frac{(mY + Z) \cdot F_1(R_e^m) + 2f(e, T_0, p_0, m)}{m(Y + 1) + Z} \\
&= \frac{(mY + Z) \cdot F_1(R_e^m) + m \cdot [2f(e, T_0, p_0, m)/m]}{(mY + Z) + m}
\end{aligned}
\tag{15}
$$

which is a weighted mean of $F_1(R_e^m)$ and $2f(e, T_0, p_0, m)/m$ with weights $mY + Z$ and $m$. Note that $Z = \sum_T \sum_{p=1}^{|T|-m+1} f(e, T, p, m) \geq 0$, and $Z = 0$ if and only if $f(e, T, p, m) = 0$ for all $T$ and $p$. If $Z = 0$, then $e$ is inactive everywhere and $F_1(R_e^m) = 0$ for any $R_e^{m\dagger}$, thus $\mathrm{SRP}(e, m) = 0$ and we can simply let $\alpha_e^m = 0$. Therefore, we assume that $Z > 0$, then both $m$ and $mY + Z$ are positive. Hence, $F_1\left(\widehat{R_e^m}\right) \geq F_1(R_e^m)$ if and only if $2f(e, T_0, p_0, m)/m \geq F_1(R_e^m)$. Equality is achieved when and only when all equalities hold.

The above result indicates that, in order to increase $F_1(R_e^m)$, for any segment satisfying $R_e^m(T, p) = 0$ and $f(e, T, p, m) \geq (m/2) \cdot F_1(R_e^m)$, we should change the routing decision to $R_e^m(T, p) = 1^{\ddagger}$, and for any segment satisfying $R_e^m(T, p) = 1$ and $f(e, T, p, m) < (m/2) \cdot F_1(R_e^m)$, we should change the routing decision to $R_e^m(T, p) = 0$. Under the case where the number of possible inputs is finite (which is the case for most LLMs due to their limited context windows), this will eventually result in a $\widehat{R_e^m}$ that activates and only activates all segments with $f(e, T, p, m) \geq (m/2) \cdot F_1\left(\widehat{R_e^m}\right)$, whose $F_1$ cannot increase further. Such $\widehat{R_e^m}$ must be unique and maximizing $F_1(R_e^m)$: Otherwise, if there exists another $\widehat{R_e^m}'$ with $F_1\left(\widehat{R_e^m}'\right) > F_1\left(\widehat{R_e^m}\right)$, then the only segments where $\widehat{R_e^m}$ and $\widehat{R_e^m}'$ disagree are the ones satisfying $(m/2) \cdot F_1\left(\widehat{R_e^m}\right) \leq f(e, T, p, m) < (m/2) \cdot F_1\left(\widehat{R_e^m}'\right)$, where $\widehat{R_e^m}(T, p) = 1$ and $\widehat{R_e^m}'(T, p) = 0$; however, changing $\widehat{R_e^m}$ on these segments to 0 should not increase $F_1\left(\widehat{R_e^m}\right)$, thus $F_1\left(\widehat{R_e^m}'\right) \leq F_1\left(\widehat{R_e^m}\right)$, a contradiction. Therefore, we can let $\alpha_{e,m} = \left\lceil F_1\left(\widehat{R_e^m}\right)\right\rceil$, which yields Equation 4. $\square$

## C  Experiment setup details

### C.1  Model architecture list

Table 4 lists the detailed architecture and configuration of all models where we conduct our experiments.

Some special notes:

- SwitchTransformers-Base-128 and NLLB-MoE-54B are encoder-decoder models that use the T5 architecture. SwitchTransformers-Base-128 has 12 encoder layers and 12 decoder layers. NLLB-MoE-54B has 24 encoder layers and 24 decoder layers.

- JetMoE-8B employs mixture-of-attention[39], which we keep intact in our experiments.

- GRIN-MoE shares the same architecture with Phi-3.5-MoE, but is trained using different methods.

- Jamba-Mini-1.6 employs a hybrid SSM-Transformer structure, yet the MoE part is identical to vanilla transformer-based MoE.

---

$\dagger$ If $Y = R_e^m(T, p) = 0$ for all $T$ and $p$, then $F_1(R_e^m)$ is undefined, which we do not concern.

$\ddagger$ When $f(e, T, p, m) = (m/2) \cdot F_1(R_e^m)$, changing $R_e^m(T, p)$ does not affect $F_1(R_e^m)$.

Table 4: Model architecture and configuration, sorted by model size. Experts: T: total; A: active; S: shared (not included in total).

| Model | # Params (B) | | # Layers | MoE Layer | # Experts | | |
|---|---|---|---|---|---|---|---|
| | Total | Active | | | T | A | S |
| PowerMoE-3B[42] | 3.30 | 0.88 | 32 | all | 40 | 8 | 0 |
| LLaMA-MoE-v1-3.5B[60] | 6.74 | 3.50 | 32 | all | 16 | 4 | 0 |
| OLMoE-1B-7B-0125[29] | 6.92 | 1.28 | 16 | all | 64 | 8 | 0 |
| SwitchTransformers-Base-128[10] | 7.42 | 0.22 | 24 | every 2 | 128 | 1 | 0 |
| LLaMA-MoE-v2-3.8B[33] | 8.03 | 3.80 | 32 | all | 8 | 2 | 0 |
| JetMoE-8B[41] | 8.52 | 2.33 | 24 | all | 8 | 2 | 0 |
| OpenMoE-8B[48] | 11.86 | 3.80 | 24 | every 6 | 32 | 2 | 1 |
| MiniCPM-MoE-8x2B[13] | 13.87 | 4.32 | 40 | all | 8 | 2 | 0 |
| Qwen1.5-MoE-A2.7B[34] | 14.32 | 2.69 | 24 | all | 60 | 4 | 4 |
| DeepSeek-V2-Lite[5] | 15.71 | 2.66 | 27 | after 1st | 64 | 6 | 2 |
| DeepSeekMoE[4] | 16.38 | 2.83 | 28 | after 1st | 64 | 6 | 2 |
| XVERSE-MoE-A4.2B[50] | 25.78 | 4.23 | 28 | all | 64 | 6 | 2 |
| Qwen3-30B-A3B[35] | 30.53 | 3.35 | 48 | all | 128 | 8 | 0 |
| Yuan2.0-M32[47] | 39.94 | 3.70 | 24 | all | 32 | 2 | 0 |
| Phi-3.5-MoE[1] | 41.87 | 6.64 | 32 | all | 16 | 2 | 0 |
| GRIN-MoE[26] | 41.87 | 6.64 | 32 | all | 16 | 2 | 0 |
| Mixtral-8x7B-v0.1[17] | 46.70 | 12.88 | 32 | all | 8 | 2 | 0 |
| Jamba-Mini-1.6[20] | 51.57 | 12.11 | 32 | every 2 | 16 | 2 | 0 |
| NLLB-MoE-54B[31] | 54.50 | 3.75 | 48 | every 4 | 128 | 2 | 0 |
| Qwen2-57B-A14B[51] | 57.41 | 14.25 | 28 | all | 64 | 8 | 8 |

Many models provide both base and post-trained (e.g., SFT) versions. We compared the local routing consistency between these versions for several models in Appendix D.4 and found no significant difference. Therefore, we always choose the base version in our main experiments.

### C.2 Data processing and input generation

We concatenate samples from each domain subset of RedPajama, cutting them into input sequences of 512 tokens (the context window size of SwitchTransformers). For each domain, we sample 2,048 input sequences, resulting in 14,336 input samples in total.

For SwitchTransformers, since the model is trained for masked language modeling, we randomly select 64 tokens from each input sequence, masking them in the original sequence as the encoder input and constructing the corresponding label sequence as the decoder input. For NLLB-MoE, as the model is trained for machine translation, we use the same sequence (with the English language token prepended) as both the encoder input and the decoder input. All other models do not need further data preprocessing, as they are decoder-only and trained for next token prediction.

## D  Additional results

### D.1  Statistical significance of local routing consistency

Due to the very high correlation between SRP and SCH, we choose SCH to represent local routing consistency, and report the 95% confidence intervals when $\rho = 2$ in Table 5. The confidence intervals are obtained by bootstrapping 1,000 times with samples from $S$.

### D.2  SRP and corpus perplexity

Figure 6 compares each model's segment routing best performance with its mean log perplexity on $S$. Most models have corpus perplexity close to each other, and we do not find a significant relation between SRP and corpus perplexity.

Table 5: Model SCH($E, m, 2$) 95% confidence interval. The decoder of SwitchTransformer does not have valid data when $m = 256$.

| Model | $m = 4$ | $m = 16$ | $m = 64$ | $m = 256$ |
|---|---|---|---|---|
| LLaMA-MoE-v2 | (99.97, 99.97) | (99.00, 99.04) | (98.15, 98.21) | (97.67, 97.76) |
| Yuan2.0 | (94.98, 95.08) | (82.36, 82.53) | (78.49, 78.69) | (76.74, 76.98) |
| Qwen3 | (93.02, 93.18) | (77.37, 77.64) | (69.22, 69.59) | (64.15, 64.64) |
| PowerMoE | (92.58, 92.67) | (79.54, 79.69) | (73.67, 73.89) | (70.94, 71.21) |
| Phi-3.5-MoE | (89.19, 89.39) | (73.91, 74.26) | (66.61, 67.07) | (62.73, 63.30) |
| GRIN-MoE | (88.25, 88.45) | (72.70, 73.05) | (65.30, 65.76) | (61.35, 61.90) |
| OLMoE | (88.72, 88.90) | (72.01, 72.34) | (64.16, 64.59) | (59.96, 60.45) |
| Mixtral-8x7B | (88.16, 88.24) | (73.75, 73.87) | (65.91, 66.07) | (61.75, 61.95) |
| MiniCPM-MoE | (88.23, 88.31) | (73.20, 73.31) | (64.96, 65.08) | (60.70, 60.84) |
| JetMoE | (85.74, 85.82) | (70.80, 70.89) | (62.86, 62.98) | (58.80, 58.95) |
| LLaMA-MoE-v1 | (80.87, 80.96) | (66.94, 67.03) | (60.05, 60.15) | (56.74, 56.86) |
| XVERSE-MoE | (77.74, 77.87) | (56.73, 56.92) | (45.53, 45.75) | (40.07, 40.33) |
| Jamba-Mini-1.6 | (77.78, 77.91) | (56.67, 56.84) | (46.00, 46.23) | (40.57, 40.87) |
| DeepSeek-V2-Lite | (77.82, 77.96) | (56.47, 56.68) | (45.37, 45.63) | (39.64, 39.95) |
| Qwen2 | (75.51, 75.59) | (55.08, 55.14) | (45.29, 45.36) | (40.90, 40.97) |
| DeepSeekMoE | (76.90, 77.05) | (54.96, 55.17) | (43.52, 43.78) | (37.68, 37.97) |
| NLLB-MoE (en) | (61.99, 62.13) | (37.00, 37.25) | (27.85, 28.14) | (23.39, 23.70) |
| (de) | (64.97, 65.18) | (44.36, 44.67) | (37.16, 37.49) | (31.33, 31.68) |
| Qwen1.5-MoE | (69.57, 69.70) | (45.56, 45.74) | (33.35, 33.56) | (27.40, 27.64) |
| OpenMoE | (64.87, 65.09) | (41.74, 42.06) | (31.76, 32.15) | (27.22, 27.67) |
| SwitchTF (en) | (54.72, 54.86) | (26.32, 26.57) | (16.71, 17.01) | (12.84, 13.16) |
| (de) | (55.16, 55.33) | (26.17, 26.47) | (16.13, 16.44) | |

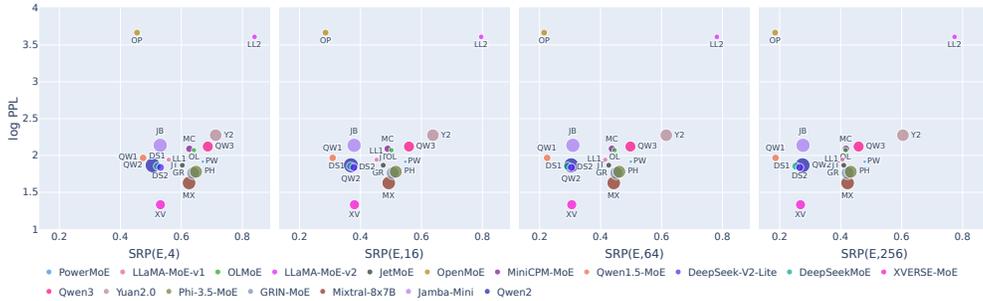

Figure 6: Model SRP on $S$, compared with mean log perplexity on $S$. Marker size represents model size.

### D.3 SRP and model architecture

Table 6 compares each model's segment routing best performance with several architecture parameters listed in Table 4 but not in Table 1. On these parameters, we do not observe significant patterns related to SRP.

### D.4 Base vs. post-trained

We selected three models—LLaMA-MoE-v1, OLMoE, and JetMoE—that have both base and post-trained versions released, and calculated SRP($E, m$) and $\rho(E, m)$ of each version. Table 7 lists the results, from which we can see that the differences of both SRP($E, m$) and $\rho(E, m)$ between models before and after post-training are not significant enough to change the degree of local routing consistency, regardless of what type of post-training (SFT, DPO, etc.) is applied. Another related fact is that Phi-MoE-3.5 and GRIN-MoE, which share the same model architecture but are trained differently, have similar local routing consistency. Both indicate that the training method may be less important than the model architecture concerning local routing consistency.

Table 6: Model segment routing best performance ($m = 16$) on $S$, compared with architecture parameters listed in Table 4 but not in Table 1.

| Model | SRP | # Params (B) | | Active Param Ratio (%) | # Layers | # Experts | |
|---|---|---|---|---|---|---|---|
| | | Total | Active | | | T | A |
| LLaMA-MoE-v2 | 79.59 | 8.03 | 3.80 | 47.36 | 32 | 8 | 2 |
| Yuan2.0 | 63.75 | 39.94 | 3.70 | 9.27 | 24 | 32 | 2 |
| Qwen3 | 55.90 | 30.53 | 3.35 | 10.98 | 48 | 128 | 8 |
| PowerMoE | 54.69 | 3.30 | 0.88 | 26.76 | 32 | 40 | 8 |
| Phi-3.5-MoE | 51.59 | 41.87 | 6.64 | 15.86 | 32 | 16 | 2 |
| GRIN-MoE | 50.55 | 41.87 | 6.64 | 15.86 | 32 | 16 | 2 |
| OLMoE | 50.17 | 6.92 | 1.28 | 18.53 | 16 | 64 | 8 |
| Mixtral-8x7B | 49.27 | 46.70 | 12.88 | 27.58 | 32 | 8 | 2 |
| MiniCPM-MoE | 48.93 | 13.87 | 4.32 | 31.13 | 40 | 8 | 2 |
| JetMoE | 47.46 | 8.52 | 2.33 | 27.36 | 24 | 8 | 2 |
| LLaMA-MoE-v1 | 45.28 | 6.74 | 3.50 | 51.85 | 32 | 16 | 4 |
| XVERSE-MoE | 38.03 | 25.78 | 4.23 | 16.39 | 28 | 64 | 6 |
| Jamba-Mini-1.6 | 37.91 | 51.57 | 12.11 | 23.48 | 32 | 16 | 2 |
| DeepSeek-V2-Lite | 37.79 | 15.71 | 2.66 | 16.94 | 27 | 64 | 6 |
| Qwen2 | 36.86 | 57.41 | 14.25 | 24.82 | 28 | 64 | 8 |
| DeepSeekMoE | 36.79 | 16.38 | 2.83 | 17.27 | 28 | 64 | 6 |
| NLLB-MoE (en) | 24.84 | 27.25 | 1.88 | 6.88 | 24 | 128 | 2 |
| (de) | 30.94 | 27.25 | 1.88 | 6.88 | 24 | 128 | 2 |
| Qwen1.5-MoE | 30.88 | 14.32 | 2.69 | 18.78 | 24 | 60 | 4 |
| OpenMoE | 28.54 | 11.86 | 3.80 | 32.07 | 24 | 32 | 2 |
| SwitchTF (en) | 18.98 | 3.71 | 0.11 | 3.02 | 12 | 128 | 1 |
| (de) | 18.90 | 3.71 | 0.11 | 3.02 | 12 | 128 | 1 |

Table 7: Segment routing best performance between models before and after post-training.

| Model | $m = 4$ | | $m = 16$ | | $m = 64$ | | $m = 256$ | |
|---|---|---|---|---|---|---|---|---|
| | SRP | $\rho(E, m)$ | SRP | $\rho(E, m)$ | SRP | $\rho(E, m)$ | SRP | $\rho(E, m)$ |
| LLaMA-MoE-v1 | 55.85 | 1.03 | 45.28 | 2.39 | 41.53 | 2.94 | 40.55 | 3.57 |
| +SFT | -0.01 | -0.00 | -0.01 | -0.00 | -0.01 | +0.00 | -0.00 | -0.00 |
| OLMoE | 64.14 | 0.99 | 50.17 | 1.05 | 44.64 | 1.20 | 41.63 | 1.22 |
| +SFT | +0.27 | +0.00 | +0.33 | +0.01 | +0.38 | +0.01 | +0.47 | +0.02 |
| +DPO | +0.32 | +0.00 | +0.41 | +0.01 | +0.47 | +0.01 | +0.58 | -0.03 |
| +Instruct | +0.29 | +0.00 | +0.35 | +0.01 | +0.40 | +0.01 | +0.50 | +0.02 |
| JetMoE | 60.31 | 1.09 | 47.46 | 2.25 | 42.69 | 2.70 | 41.01 | 3.18 |
| +SFT | -0.18 | -0.00 | -0.12 | +0.01 | -0.10 | +0.01 | -0.06 | +0.02 |
| +Chat | -0.19 | -0.00 | -0.13 | +0.01 | -0.11 | +0.01 | -0.07 | +0.03 |

### D.5 SRP per segment position

To determine whether the segment position $p$ can affect the segment routing best performance, we calculate $SRP(E, m)$ on each segment position by summarizing statistics of all segments that share the same position. Figure 7 illustrates this position-wise $SRP(E, m)$ at each possible segment position. Most models have nearly constant $SRP(E, m)$ at every position except $p = 0$, where many models activate specialized experts to handle the beginning of the input sequence. This stability of local routing consistency across input positions allows us to use segments from all positions to calculate $SRP(E, m)$, and apply conclusions based on $SRP(E, m)$ to any segment of the input (except the very first one).

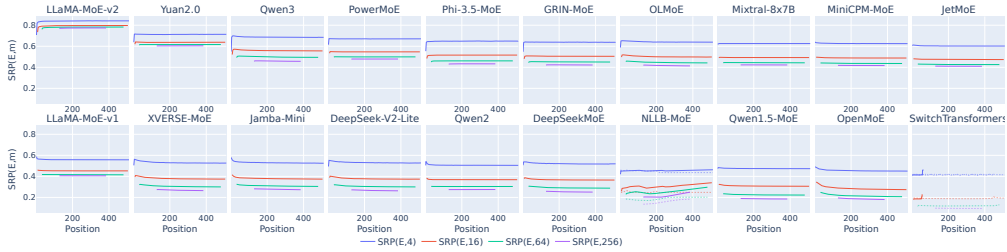

Figure 7: Position-wise segment routing best performance on $S$ of each model. For encoder-decoder models, dotted lines show the encoder $SRP(E, m)$ and solid lines show the decoder ones.

### D.6 SRP across domains

To verify whether local routing consistency is transitive across different domains, we calculate the correlation of expert segment routing best performance between pair-wise domains and demonstrate it in Figure 8. We also compute the correlation of expert activation frequency between pair-wise domains, results illustrated in Figure 9. By comparing corresponding heapmaps, we can see that local routing consistency is nearly always positively correlated, even between distant domains on which the experts' activation frequencies are negatively correlated. This means that local routing consistency is transitive; domain-specialized experts with high local routing consistency in one domain tend to exhibit it in any other domain. We also found that some models (e.g., LLaMA-MoE-v2 and Qwen2) do not show a significant difference between domains, which is aligned with the results in Section 4.2.

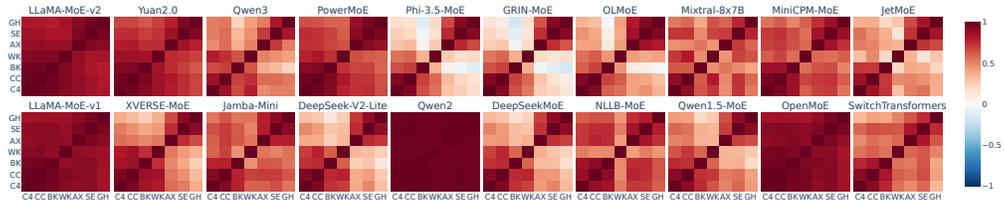

Figure 8: Correlation between domain-wise expert SRP of each model. CC: CommonCrawl; BK: Books; WK: Wikipedia; AX: ArXiv; SE: StackExchange; GH: GitHub.

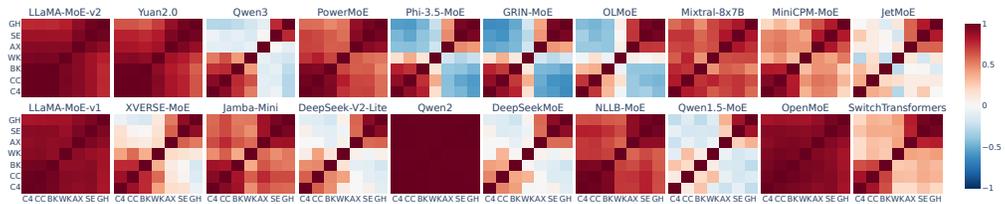

Figure 9: Correlation between the domain-wise expert activation frequency of each model. CC: CommonCrawl; BK: Books; WK: Wikipedia; AX: ArXiv; SE: StackExchange; GH: GitHub.

### D.7 Layer level results

Figure 10 illustrates each model's layer-wise SRP. Most models have peak SRPs among middle layers, while some (e.g., Yuan2.0 and MiniCPM) have another peak at the last layer. We conjecture that middle layers are less tied to input/output tokens and thus more sensitive to the general topic, and the final layers process highly abstract information that is also more related to the overall topic. Both encourage routers to select similar experts within a local segment that share the same topic across tokens. PowerMoE and Qwen2 have another peak on layer 2 due to expert imbalance. Appendix D.8 gives a clear view on this.

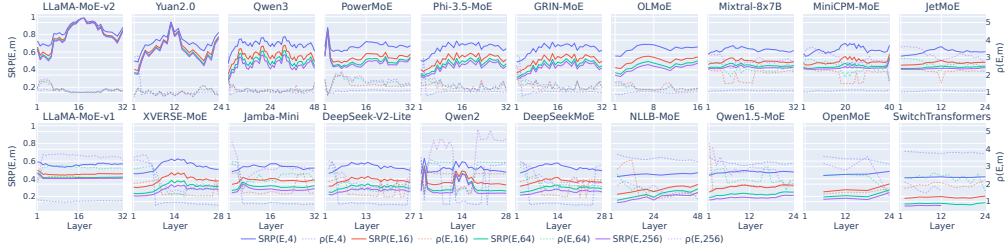

Figure 10: Layer-wise SRP on $S$ of each model. Solid lines show $\mathrm{SRP}(E, m)$ while dotted lines show corresponding $\rho(E, m)$.

We also calculated layer-wise segment cache best hit rate, results demonstrated in Figure 11. The patterns are the same as SRP, indicating a high correlation between the two metrics.

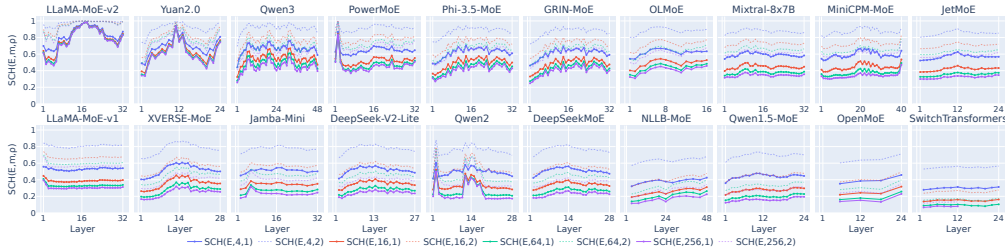

Figure 11: Layer-wise segment cache best hit rate on $S$ of each model. Solid lines show $\mathrm{SCH}(E, m, 1)$ while dotted lines show corresponding $\mathrm{SCH}(E, m, 2)$.

### D.8 Expert level results

We demonstrate expert-wise segment routing best performance against activation frequency in Figure 12. LLaMA-MoE-v2, Yuan2.0, and PowerMoE have experts with very high activation frequency. These experts naturally have very high local routing consistency and contribute to these models' high model-level local routing consistency. The imbalanced experts of PowerMoE mainly belong to layer 2, which also explains the observation in Section D.7.

Furthermore, Figures 13, 14, 15 and 16 compares SRP with domain and vocabulary specializations. The plots are aligned with the conclusion of Section 4.2 that when the model exhibits domain specialization, domain-specialized experts contribute more to overall local routing consistency than vocabulary-specialized experts.

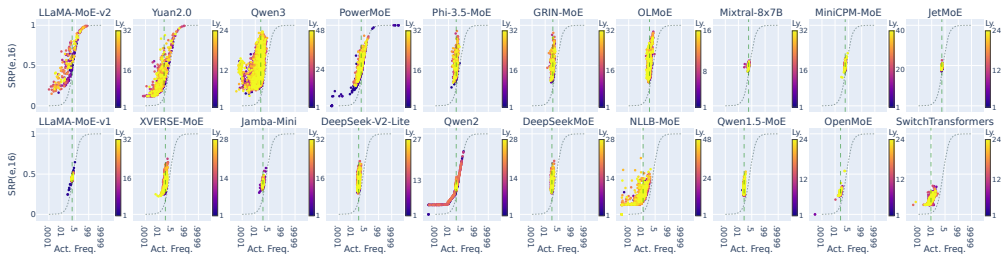

Figure 12: Per-expert activate frequency vs. SRP. The x-axis is stretched to show experts with very low or high activation frequency. Gray dashed lines indicate the theoretical lower bound of SRP at different activation frequencies. Green dashed lines show the expected activation frequency of experts from each model.

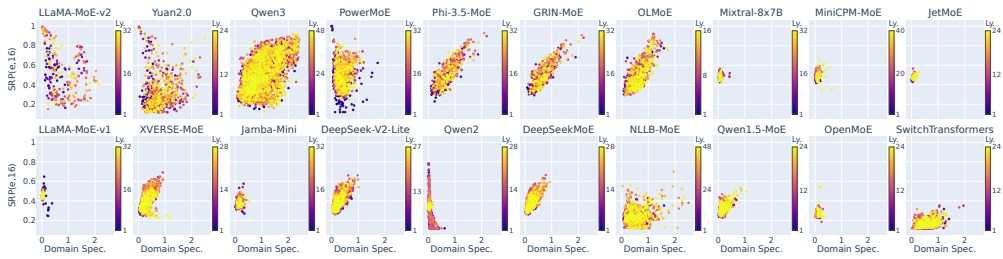

Figure 13: Per-expert domain specialization vs. SRP

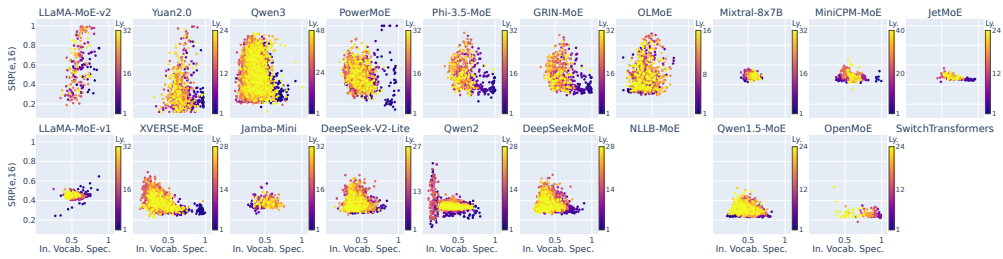

Figure 14: Per-expert input vocabulary specialization vs. SRP. Encoder-decoder models are not involved due to different input formats from other decoder-only models.

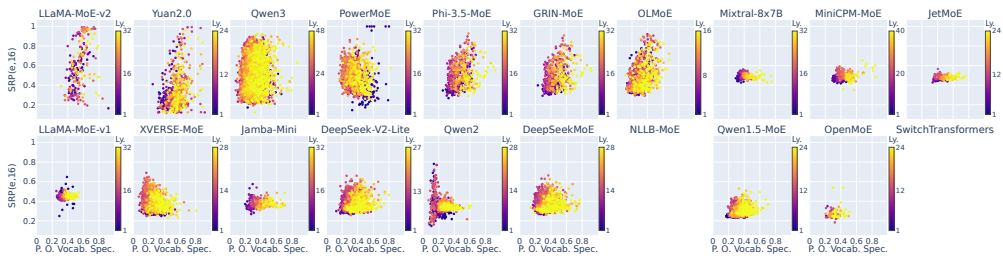

Figure 15: Per-expert predicted output vocabulary specialization vs. SRP. Encoder-decoder models are not involved due to different input formats from other decoder-only models.

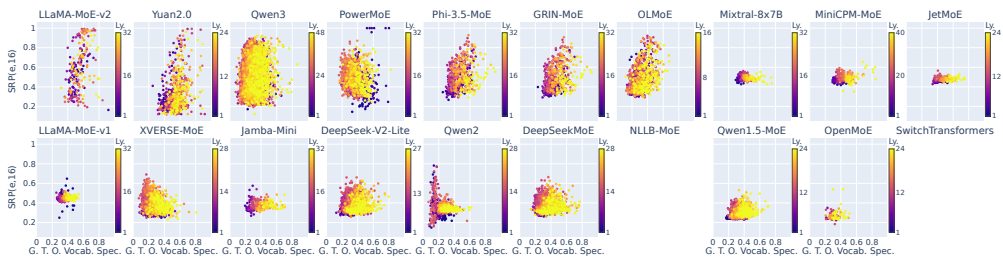

Figure 16: Per-expert ground-truth output vocabulary specialization vs. segment routing best performance. Encoder-decoder models are not involved due to different input formats from other decoder-only models.